\documentclass{article}

\usepackage{microtype}
\usepackage{graphicx}
\usepackage{subfigure}
\usepackage{booktabs} 
\usepackage{amsmath}
\usepackage{enumitem}  
\usepackage{tikz}
\usetikzlibrary{positioning, arrows.meta, shapes.geometric, fit, backgrounds, calc}
\usepackage{booktabs}
\usepackage{algorithmic}
\usepackage{amsfonts}

\usepackage{hyperref}

\usepackage{algorithm}

\usepackage[accepted]{mlsys2025}

\makeatletter
\renewcommand{\mlsys@appearing}{Preprint.}
\makeatother

\mlsystitlerunning{Follow the Features}

\begin{document}

\twocolumn[
\mlsystitle{Follow the Features: SAE Feature-Guided Data-Efficient Instruction-Tuning of Large Language Models}




\begin{mlsysauthorlist}
\mlsysauthor{Ryan Swift}{usc}
\mlsysauthor{Konstantinos Psounis}{usc}
\end{mlsysauthorlist}

\mlsysaffiliation{usc}{Thomas Lord Department of Computer Science, University of Southern California, Los Angeles, CA, USA}

\mlsyscorrespondingauthor{Ryan Swift}{ryanswif@usc.edu}

\mlsyskeywords{Machine Learning, interpretabililty, finetuning, fine-tuning, instruction-tuning, sparse autoencoders, data-efficiency}

\vskip 0.3in

\begin{abstract}
Selecting small, effective instruction-tuning datasets requires identifying which aspects of the original data to retain. We introduce Follow the Features (FtF), a data-selection method that uses sparse autoencoder (SAE) features to characterize training data and guide subset construction. FtF ranks features using dataset-level activation statistics, scores samples by their expression of highly ranked features, and incorporates feature resampling to encourage diversity. The LLM–SAE pair used for selection is independent of the model being fine-tuned, enabling reuse across target models. We evaluate FtF on two model–dataset settings at a training budget of 100 samples with a fixed sequence length of 256 tokens. FtF achieves performance competitive with GRAPE and S2L, with relative rankings varying across settings. On Qwen 3 8B trained on Alpaca, FtF exceeds full-dataset training by 1.59 percentage points in the harmonic mean of an AlpacaEval-based win rate and Open LLM Leaderboard performance, although full-dataset training retains a higher Leaderboard score. These results support SAE feature statistics as a useful signal for data-efficient instruction tuning.
\end{abstract}
]



\printAffiliationsAndNotice{}

\section{Introduction}
\label{s:intro}

Instruction tuning adapts pretrained large language models (LLMs) to follow instructions and provide useful responses \citep{ouyang_training_2022, taori_alpaca_2023, zheng_judging_2023}. Although large instruction datasets can support this process, training on every available example is not always necessary. Data selection offers an opportunity to reduce fine-tuning costs by identifying smaller subsets that retain useful training signals. The central challenge is deciding which properties of the original dataset to preserve under a limited training budget.

Existing approaches address this challenge through different notions of sample value. GRAPE favors examples that align with the target model's existing distribution \citep{zhang_best_2025}, while SmallToLarge (S2L) uses the training dynamics of a smaller proxy model to identify informative examples \citep{yang_smalltolarge_2024}. A complementary approach is to examine the features expressed by the data. Sparse autoencoders (SAEs) decompose LLM activations into sparse feature representations, many of which have been associated with recognizable linguistic and semantic patterns \citep{bricken_towards_2023, gao_scaling_2024, templeton_scaling_2024}. These representations provide a basis for characterizing datasets through the internal responses they elicit from a model.

Feature-based selection raises a question beyond whether a feature is present: how strongly and consistently is it expressed across a dataset? Prior work uses SAE feature coverage to encourage diversity in instruction-tuning subsets \citep{yang_diversity-driven_2025}. However, feature presence alone does not distinguish weak from strong expression or account for differences in activation frequency and variability. We investigate whether these additional statistics provide a useful signal for selecting training examples. Our working hypothesis is that samples expressing a dataset's prominent features can concentrate useful training signals into a small subset. We treat this as a hypothesis to evaluate through downstream performance, rather than assuming that prominent features necessarily correspond to the concepts most valuable for learning.

We introduce \textit{Follow the Features} (FtF), a data-selection method that constructs instruction-tuning subsets using dataset-level SAE feature statistics. FtF first processes the source dataset with an LLM--SAE \textit{filter pair} and ranks features using their mean activation, decoder-vector norm, and conditional activation entropy. It then scores samples according to their expression of the top-ranked features and selects a subset of a specified size. A feature-resampling step introduces additional features to encourage diversity beyond the highest-ranked set. The filter pair is independent of the model being fine-tuned, allowing the same feature representation to guide selection for different target models.

We evaluate FtF at a training budget of 100 samples with a fixed sequence length of 256 tokens, comparing it with random selection, GRAPE, S2L, and full-dataset training. Our main experiments consider Llama 3.1 8B trained on UltraChat 200K and Qwen 3 8B trained on Alpaca (Cleaned). We assess both instruction-following preferences through an AlpacaEval-based evaluation and benchmark performance through Open LLM Leaderboard v2. FtF is competitive with the evaluated selection methods, although their relative rankings vary across the two settings. For Qwen trained on Alpaca, FtF exceeds full-dataset training by 1.59 percentage points in the harmonic mean of the two evaluation scores. Full-dataset training retains the highest Leaderboard aggregate in both settings, highlighting a tradeoff between the measured outcomes.

Our contributions are threefold. First, we introduce a transferable data-selection procedure that uses the magnitude and distribution of SAE activations to rank features and training samples. Second, we evaluate this procedure against established selection methods under a common, tightly constrained training-data budget. Third, we analyze its design choices, the use of a smaller filter model, and the agreement of feature rankings across datasets. Together, these experiments investigate the utility of SAE feature statistics for data-efficient instruction tuning while identifying open questions about the interpretation of selected features and their relationship to downstream learning.

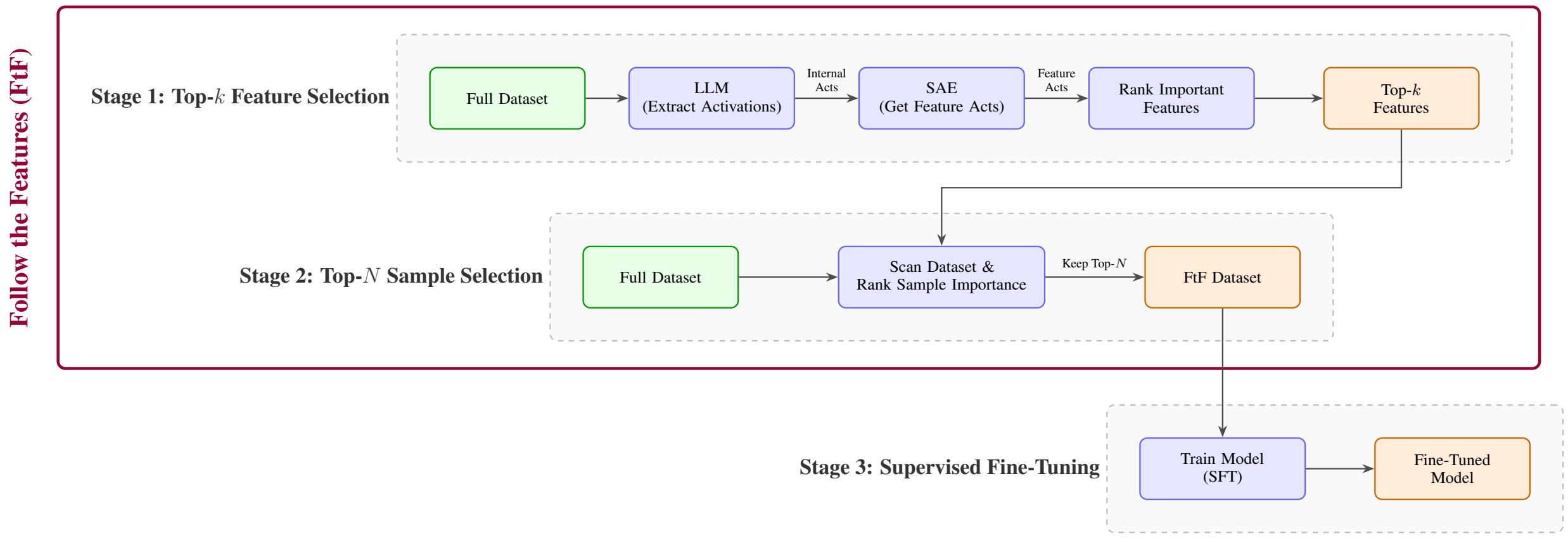
\begin{figure*}[htbp]
    \centering
    \resizebox{0.95\textwidth}{!}{%
    \begin{tikzpicture}[
        node distance = 1.5cm and 2cm,
        process/.style = {
            rectangle,
            draw=blue!60,
            thick,
            rounded corners,
            fill=blue!10,
            text width=3cm,
            align=center,
            minimum height=1.2cm,
            font=\small
        },
        data/.style = {
            rectangle,
            draw=green!60!black,
            thick,
            rounded corners,
            fill=green!10,
            text width=2.8cm,
            align=center,
            minimum height=1.2cm,
            font=\small
        },
        deliverable/.style = {
            rectangle,
            draw=orange!80!black,
            thick,
            rounded corners,
            fill=orange!15,
            text width=2.8cm,
            align=center,
            minimum height=1.2cm,
            font=\small
        },
        arrow/.style = {-Stealth, thick, draw=black!70},
        stagebox/.style = {
            draw=gray!50,
            dashed,
            thick,
            inner sep=18pt,
            rounded corners,
            fill=gray!5
        }
    ]

    \def\stageOneY{0}
    \def\stageTwoY{-3.5}
    \def\stageThreeY{-7.25}

    \def\featureRouteY{-1.75}

    \node[data] (inputs) at (0, \stageOneY) {Full Dataset};
    \node[process] (llm) at (4, \stageOneY)
        {LLM\\(Extract Activations)};
    \node[process] (sae) at (8.5, \stageOneY)
        {SAE\\(Get Feature Acts)};
    \node[process] (rank) at (13, \stageOneY)
        {Rank Important\\Features};
    \node[deliverable] (topk) at (17.5, \stageOneY)
        {Top-$k$\\Features};

    \draw[arrow] (inputs) -- (llm);
    \draw[arrow] (llm) --
        node[above, font=\scriptsize, align=center]
        {Internal\\Acts}
        (sae);
    \draw[arrow] (sae) --
        node[above, font=\scriptsize, align=center]
        {Feature\\Acts}
        (rank);
    \draw[arrow] (rank) -- (topk);

    \node[data] (dataset) at (3, \stageTwoY)
        {Full Dataset};
    \node[process, text width=3.8cm] (filter) at (8.5, \stageTwoY)
        {Scan Dataset \&\\Rank Sample Importance};
    \node[deliverable] (subset) at (14, \stageTwoY)
        {FtF Dataset};

    \draw[arrow] (dataset) -- (filter);

    \draw[arrow]
        (topk.south) --
        (17.5, \featureRouteY) --
        (8.5, \featureRouteY) --
        (filter.north);

    \draw[arrow] (filter) --
        node[above, font=\scriptsize, align=center]
        {Keep Top-$N$}
        (subset);

    \node[process] (train) at (14, \stageThreeY)
        {Train Model\\(SFT)};
    \node[deliverable] (final) at (18.5, \stageThreeY)
        {Fine-Tuned\\Model};

    \draw[arrow] (subset.south) -- (train.north);
    \draw[arrow] (train) -- (final);

    \begin{scope}[on background layer]
        \node[
            stagebox,
            fit=(inputs) (topk),
            label={
                [name=lbl1, anchor=east,
                 font=\bfseries\large, text=black!80]
                left:Stage 1: Top-$k$ Feature Selection
            }
        ] (stage1) {};

        \node[
            stagebox,
            fit=(dataset) (filter) (subset),
            label={
                [name=lbl2, anchor=east,
                 font=\bfseries\large, text=black!80]
                left:Stage 2: Top-$N$ Sample Selection
            }
        ] (stage2) {};

        \node[
            stagebox,
            fit=(train) (final),
            label={
                [anchor=east,
                 font=\bfseries\large, text=black!80]
                left:Stage 3: Supervised Fine-Tuning
            }
        ] (stage3) {};

        \node[
            draw=purple!80!black,
            ultra thick,
            rounded corners,
            inner sep=15pt,
            fit=(stage1) (stage2) (lbl1) (lbl2)
        ] (pruning) {};

        \node[
            font=\bfseries\Large,
            text=purple!80!black,
            rotate=90
        ] at ([xshift=-20pt]pruning.west)
        {Follow the Features (FtF)};
    \end{scope}

    \end{tikzpicture}%
    }
    \caption{
        The generalized Follow the Features (FtF) pipeline. FtF encompasses
        top-$k$ feature selection and top-$N$ sample selection to
        produce a refined data subset of $N$ samples, which is then
        used for instruction-tuning via Supervised Fine-Tuning (SFT).
        The set of top-$k$ features, the pruned dataset, and the trained
        model are the deliverable items produced by our pipeline. We
        evaluate the quality of FtF on the basis of the task performance of
        the trained model.
    }
    \label{fig:ftf_pipeline}
\end{figure*}

\section{Preliminary: Sparse Autoencoders}
\label{s:prelims}

Sparse autoencoders are a type of self-supervised neural network where inputs are encoded ($E$) into a latent space and then decoded ($D$) into a reconstruction of the inputs. Sparsity is often enforced by training with an L1 penalty ($\alpha$) on the latent activations, as in Equation \ref{eq:sae_loss}.
\begin{equation}
\label{eq:sae_loss}
\mathcal{L} = \underbrace{\|x - D(E(x))\|^2}_{\text{Reconstruction}} + \alpha \underbrace{\sum_{i=1}^{N} |E(x)_i|}_{\text{Sparsity}}
\end{equation}
SAEs have been a powerful tool for working toward a human-interpretable understanding of LLMs. 
SAEs decompose the dense LLM activations into a sparse, overcomplete latent space where the reconstructed inputs are an affine transformation of the latents. These latents have been mapped to individual semantic concepts which demonstrate the important property of semantic causality \citep{bricken_towards_2023, gao_scaling_2024, templeton_scaling_2024, marks_sparse_2025}. Figure \ref{fig:autoencoder_latent_multiplication} illustrates how each individual latent contributes to reconstruction through its activation and decoder vector.

\section{Background and Related Work}
\label{s:bg-rw}
\subsection{Feature-informed Data Selection}
\citet{yang_diversity-driven_2025} also leverages SAEs and interpretability for IT data selection. Their approach focuses on measuring feature diversity, but treats all active (i.e. nonzero) features equally.
They form data subsets by merging in samples containing features not yet present in the subset. The core contributions of this work was producing an explanation of the advantage of training on long sequences \citep{zhao_long_2024} via a feature-presence analysis of the training data.

The key difference in our core methodologies is that \citet{yang_diversity-driven_2025} does not make use of the additional information of feature activations or decoder norms, while FtF does. Feature activations can intuitively be understood as the strength of expression of the feature, and are mathematically tied to the SAE decoder norms. Rather than solely capture feature presence, FtF builds a more detailed profile of the features in a dataset by capturing strength of feature expression.

\subsection{Data-Efficient Instruction-Tuning}

\subsubsection{Alignment with Pretraining Distribution}
This paradigm suggests that data similar to the pretraining distribution maximizes training benefits. The \textbf{GRAPE} framework \citep{zhang_best_2025} selects samples that minimize distribution shift during supervised finetuning (SFT). By aligning IT samples with the model's existing policy distribution, GRAPE reduces the difficulty of adaptation and improves data efficiency.

\subsubsection{Small Models (Superfiltering)}
Small-to-large approaches, or \textit{superfiltering}, use smaller models to curate data for larger ones. \citet{li_superfiltering_2024} found that metrics like IFD are consistent across scales, demonstrating that GPT-2-based filtering can yield superior 7B and 13B models. \textbf{SmallToLarge (S2L)} \citep{yang_smalltolarge_2024} concretely implements this by using \textit{loss trajectories}; samples that maintain high loss over multiple epochs on a small model are flagged as high-difficulty, high-value targets for larger models.

\subsection{Human Intervention and Interpretability}
Beyond automated selection, human-in-the-loop interpretability offers practical and meaningful improvements in data processing. \citet{marks_sparse_2025} used SAEs to identify and interpret model circuits, building a semantic map of model processing. Their \textbf{SHIFT} method allows humans to ablate circuits associated with task-irrelevant or spurious features, improving performance by refining the model's focus during inference.

\section{Methodology}
\label{s:methodology}

\subsection{Intuition and Motivation}
\label{ss:intuition}
LLMs exhibit a remarkable ability to encode and organize semantic information and relationships in their internal representation spaces \cite{park_geometry_2025, gurnee_language_2024, park_linear_2024, gammelgaard_large_2023}. This challenges one of the most famous criticisms of LLMs as being nothing more than "stochastic parrots" who simply learn statistical patterns in the data without an understanding of the true semantic meaning of the text being generated \cite{bender_dangers_2021}. 

Thus, if the evidence suggests that models are learning to understand semantics, what if we look at model training through the lens of concepts and features? Given a dataset known to produce high-quality instruction-following models, we can apply the feature lens to identify the important features we want the model to learn.

The core mechanism of Follow the Features (FtF) decomposes raw training samples into their corresponding SAE features, as identified by a given LLM-SAE model pair. We hypothesize that there is some SAE feature "kernel" in a given dataset which represents the most "important" collection of features within the dataset. Furthermore, we hypothesize that a model which can capture and understand the "kernel" will have learned a meaningful proportion of the information contained within the training data. 

Our hypotheses lead us to expect that FtF can be used to create a smaller, filtered dataset which produces high quality models at a reduced training budget.
We expect this because we can create a data subset which focuses on capturing the feature "kernel" and excludes less "important" features.

In the rest of this section, we concretely define our abstract notions of a feature "kernel", and feature and sample "importance". We describe the core of the FtF methodology, as well as the application to data filtering.

\subsection{Definitions}

\subsubsection{Feature Importance}
We concretely define our notion of feature importance in Equation \ref{eq:importance_w_ent}. 

\begin{equation}
    \text{feature\_importance}_j = \frac{\mu_j \times \|w_j\|}{H(F_j \mid F_j > 0)}
    \label{eq:importance_w_ent}
\end{equation}
where,
\begin{itemize}[nosep]
    \item $j$ is the index of the feature
    \item $\mu_j$ is the mean activation of feature $j$ over the dataset
    \item $\|w_j\|$ is the norm of the SAE decoder vector for feature $j$
    \item $F_j$ is the set of all activations of feature $j$
\end{itemize}

Intuitively, it captures 3 key aspects of the feature: frequency, strength, and stability. Frequency is simple and intuitive: if the feature is common in the dataset, then it is probably important. Feature strength is measured as the influence of the feature on the reconstructed activations (i.e. the feature's activation magnitude and decoder norm). Stability is a measure of activation consistency. Certainly all features will activate with variations in magnitude. However, high activation fluctuations produce signals more difficult for models to learn.

Frequency and strength are captured in the mean feature activation: mild-strength but common features are similarly valued as somewhat less common but very strong features. The decoder norms also capture a complementary component of feature strength, as the value of the norm has a direct influence on the value of the activation. Stability is captured through the conditional entropy of nonzero activations. That is, the stability of feature activations when they are actually active.

This heuristic allows us to capture features with less frequent, higher activations as well as features with more frequent, lower activations. Features with overall strong, consistent activations are clearly preferred, and features with overall weak and inconsistent activations are disfavored.

\subsubsection{Feature Kernel: Top-$k$ Features}
We define the feature kernel as the top-$k$ most important features in a dataset, with $k$ being a tunable parameter. Increasing $k$ creates a kernel with more feature diversity, but risks diluting the signals from higher importance features. Decreasing $k$ creates a more focused kernel, where every feature included is of very high importance, but can result in a failure to capture meaningful features that happened to fall outside of the absolute highest tier of importance. Thus, tuning $k$ is a tradeoff between focus on the most important features and capturing feature diversity.

\subsubsection{Sample Importance}
Sample importance is a function of the top-$k$ most important features. It is a measure of the content and expression of the most important features in a sample, and represents how well a particular sample captures the feature kernel of the dataset. Our goal in data filtering is to approximate the feature content of original dataset using fewer samples. Thus, samples which have high expression of many of the top-$k$ features are interpreted as better approximations of the original dataset than samples with lower expression of fewer top features.

\subsection{Core Mechanism}

\begin{figure}[tbp]
    \centering
    \resizebox{0.75\columnwidth}{!}{
        \begin{tikzpicture}[
            token/.style={draw, rectangle, minimum width=1.6cm, minimum height=0.6cm, fill=white, rounded corners, font=\sffamily\small},
            dense_vec/.style={draw, rectangle, minimum width=1.6cm, minimum height=0.6cm, fill=gray!15, font=\sffamily\small},
            sparse_vec/.style={draw, rectangle, minimum width=1.6cm, minimum height=0.6cm, fill=orange!20, font=\sffamily\small, rounded corners},
            llm_block/.style={draw=blue!80, rectangle, minimum width=8.5cm, minimum height=0.7cm, fill=blue!10, rounded corners, font=\sffamily\small\bfseries, thick, align=center},
            sae_block/.style={draw=green!60!black, rectangle, minimum width=8.5cm, minimum height=0.7cm, fill=green!10, rounded corners, font=\sffamily\small\bfseries, thick, align=center},
            agg_block/.style={draw=purple!70, rectangle, minimum width=8.5cm, minimum height=0.7cm, fill=purple!10, rounded corners, font=\sffamily\small\bfseries, thick, align=center},
            final_vec/.style={draw=orange!80!black, rectangle, minimum width=4cm, minimum height=0.7cm, fill=orange!40, font=\sffamily\small\bfseries, rounded corners, thick},
            feat_agg_block/.style={draw=teal!70, rectangle, minimum width=8.5cm, minimum height=0.7cm, fill=teal!10, rounded corners, font=\sffamily\small\bfseries, thick, align=center},
            scalar_vec/.style={draw=red!80!black, rectangle, minimum width=4cm, minimum height=0.7cm, fill=red!20, font=\sffamily\small\bfseries, rounded corners, thick},
            arrow/.style={-{Stealth[scale=0.9]}, thick, draw=black!80},
            extract_arrow/.style={-{Stealth[scale=0.9]}, thick, draw=red!80}
        ]
        
        \node (t1) at (-3.2, 3.5) [token] {$t_1$};
        \node (t2) at (-1.1, 3.5) [token] {$t_2$};
        \node (tdots) at (1.0, 3.5) {$\cdots$};
        \node (tT) at (3.2, 3.5) [token] {$t_T$};
        \node at (0, 4.2) [font=\sffamily\small\bfseries] {Input Sequence of Length $T$};
        
        \node (llm) at (0, 2.2) [llm_block] {Large Language Model};
        \node (x1) at (-3.2, 0.9) [dense_vec] {$x_1 \in \mathbb{R}^d$};
        \node (x2) at (-1.1, 0.9) [dense_vec] {$x_2 \in \mathbb{R}^d$};
        \node (xdots) at (1.0, 0.9) {$\cdots$};
        \node (xT) at (3.2, 0.9) [dense_vec] {$x_T \in \mathbb{R}^d$};
        
        \node (sae) at (0, -0.2) [sae_block] {SAE Encoder};
        \node (v1) at (-3.2, -1.3) [sparse_vec] {$v_1 \in \mathbb{R}^n$};
        \node (v2) at (-1.1, -1.3) [sparse_vec] {$v_2 \in \mathbb{R}^n$};
        \node (vdots) at (1.0, -1.3) {$\cdots$};
        \node (vT) at (3.2, -1.3) [sparse_vec] {$v_T \in \mathbb{R}^n$};
        
        \node (agg) at (0, -2.6) [agg_block] {Sequence Aggregation ($v_{feat} = Agg_{seq}(v_1, \dots, v_T)$)};
        \node (vagg) at (0, -3.8) [final_vec] {$v_{feat} \in \mathbb{R}^n$};
        
        \node (featagg) at (0, -5.0) [feat_agg_block] {Feature Aggregation ($s = Agg_{feat}(v_{feat})$)};
        \node (sampleimp) at (0, -6.2) [scalar_vec] {Sample Importance $s \in \mathbb{R}$};
        
        \draw[arrow] (t1) -- (t1 |- llm.north);
        \draw[arrow] (t2) -- (t2 |- llm.north);
        \draw[arrow] (tT) -- (tT |- llm.north);
        \draw[extract_arrow] (x1 |- llm.south) -- (x1) node[midway, right, font=\tiny, text=red!80] {Extract};
        \draw[extract_arrow] (x2 |- llm.south) -- (x2);
        \draw[extract_arrow] (xT |- llm.south) -- (xT);
        \draw[arrow] (x1) -- (x1 |- sae.north);
        \draw[arrow] (x2) -- (x2 |- sae.north);
        \draw[arrow] (xT) -- (xT |- sae.north);
        \draw[arrow] (v1 |- sae.south) -- (v1);
        \draw[arrow] (v2 |- sae.south) -- (v2);
        \draw[arrow] (vT |- sae.south) -- (vT);
        \draw[arrow] (v1) -- (v1 |- agg.north);
        \draw[arrow] (v2) -- (v2 |- agg.north);
        \draw[arrow] (vT) -- (vT |- agg.north);
        \draw[arrow, thick] (agg.south) -- (vagg.north);
        \draw[arrow, thick] (vagg.south) -- (featagg.north);
        \draw[arrow, thick] (featagg.south) -- (sampleimp.north);
        
        \begin{scope}[on background layer]
            \node[fit=(x1)(sae)(sampleimp), draw=black!20, dashed, thick, fill=gray!5, rounded corners, inner sep=0.3cm] (bg) {};
        \end{scope}
        \end{tikzpicture}
    }
    \caption{Illustration of Algorithm \ref{alg:top-n-data}'s pipeline for assigning sample importance scores}
    \label{fig:sae_sequence_aggregation}
\end{figure}
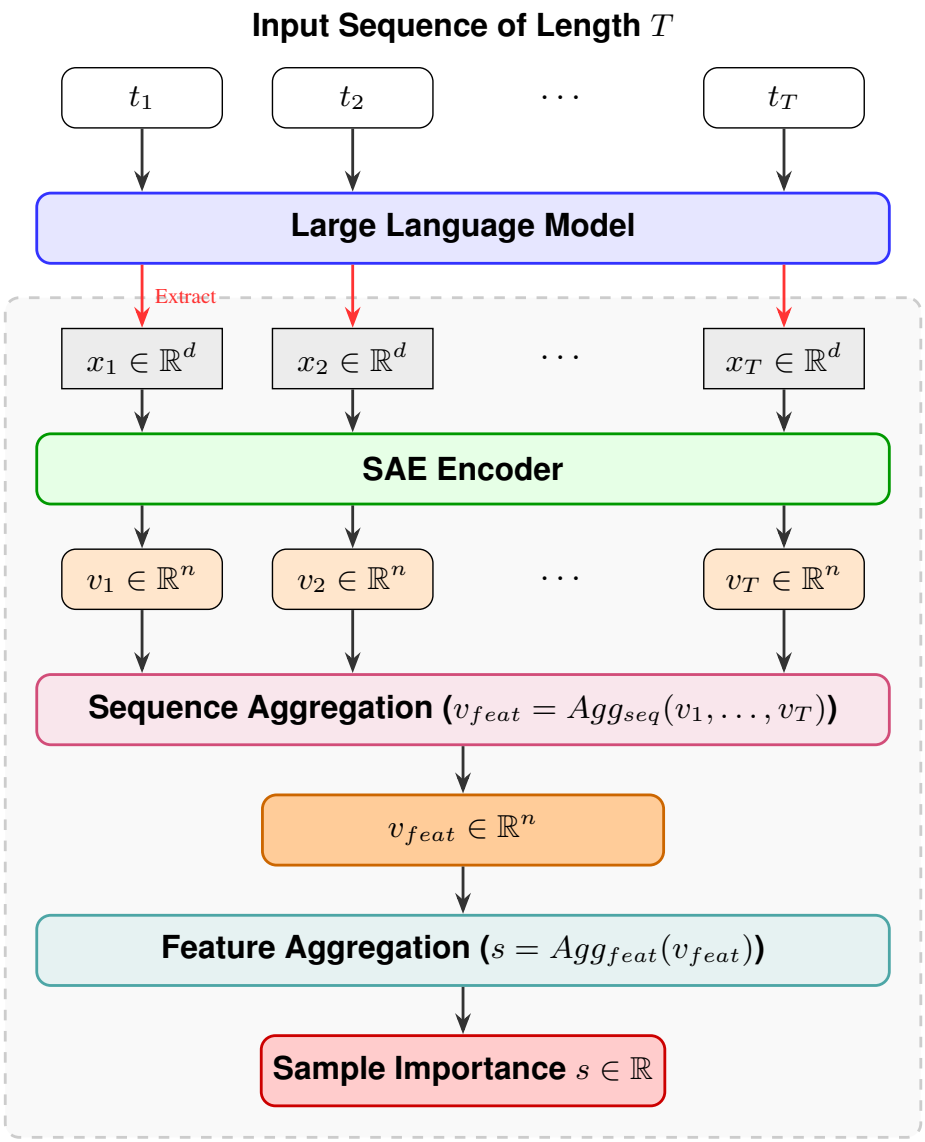

To arrive at a feature decomposition of the data, we use an LLM-SAE pair, which we refer to as the \textit{filter pair}. Since the activations in the middle layers of LLMs are known to produce rich representations which balance information compression and signal preservation \citep{skean_layer_2025}, we extract LLM activations from the residual stream after the middle layer for input into the paired SAE. 

We follow Algorithm \ref{alg:top-k-feats} to produce feature activations for each data sample, determine the importance of each feature over the full dataset, and identify our feature kernel of the top-$k$ most important features. We then follow Algorithm \ref{alg:top-n-data} to rank data samples on the basis of their expression of the top-$k$ important features. The feature activations for a given data sample are filtered to only keep those of the top-$k$ features, then an aggregation is performed over every position in the sample's token sequence ($Agg_{seq}$) to produce a vector representing the feature content of the entire data sample ($v_{feat}$). Finally, an aggregation over this feature vector ($Agg_{feat}$) is performed to arrive at the final sample importance scores. The top-$N$ most important samples are then returned to produce a dataset of a desired size.

In our main experiments described in Section \ref{experiments}, we use last-token-pooling for the sequence aggregation\footnote{In Section \ref{ss:seq-agg} we explore using mean-pooling for the sequence aggregation} (i.e. only looking at the last token position) because it is the only position which represents the full feature content of the entire sequence. We then use mean-pooling\footnote{Given that the number of features is fixed at $k$, mean-pooling is equivalent to sum-pooling (up to the constant factor $k$) in our feature aggregation} for the feature aggregation.

\begin{algorithm}
\caption{Top-$k$ Feature Selection}
\begin{algorithmic}
\REQUIRE Dataset $X$, LLM $M$, SAE $S$, output size $k$

\FOR{each sequence $x_i \in X$}
    \STATE $a_i \gets \text{ExtractActivations}(M, x_i)$
    \STATE $f_i \gets S.\text{encode}(a_i)$
\ENDFOR

\STATE $F \gets$ feature activations for all $x_i \in X$
\STATE $F_j \gets$ all activations for feature $j$
\STATE $\mu_j \gets$ mean activation for feature $j$
\STATE $w_j \gets$ decoder weight vector for feature $j$

\FOR{each feature $j$}
    \STATE \qquad $\text{feature\_importance}_j = \frac{\mu_j \times \|w_j\|}{H(F_j \mid F_j > 0)}$
\ENDFOR

\STATE \textbf{return:} The top $k$ most important features, $\mathcal{K}$
\end{algorithmic}
\label{alg:top-k-feats}
\end{algorithm}

\vspace{2em}

\begin{algorithm}
\caption{Top-$N$ Sample Selection}
\begin{algorithmic}
\REQUIRE Dataset $X$, LLM $M$, SAE $S$, top-$k$ features $\mathcal{K}$, sample size $N$, $\text{Agg}_{seq}()$, $\text{Agg}_{feat}()$

\FOR{each sequence $x_i \in X$}
    \STATE $a_i \gets \text{ExtractActivations}(M, x_i)$ 
    \STATE $f_i \gets S.\text{encode}(a_i)$
    \STATE $f_{\mathcal{K}, i} \gets$ $f_i$ filtered for features in $\mathcal{K}$
    
    \STATE $v_{feat} \gets \text{Agg}_{seq}(F_{\mathcal{K}, i})$ 
    
    \STATE $\text{sample\_importance}_i \gets \text{Agg}_{feat}(v_{feat})$ 
\ENDFOR

\STATE \textbf{return:} The top $N$ most important samples  
\end{algorithmic}
\label{alg:top-n-data}
\end{algorithm}

\subsection{Application to Dataset Filtering}
To achieve data-efficient finetuning, we build an extension of the core mechanism to perform data filtering. First, we set $N < \lvert X \rvert$, meaning that the number of samples returned from Algorithm 2 is less than the size of the original dataset. 

Once we acquire $\mathcal{K}$, our set of top features, we apply a diversity resampling according to a parameter, $\alpha$. We take the bottom $ k \times \alpha$ features in $\mathcal{K}$ and uniformly sample from the remaining SAE features. This resampling further promotes coverage of the feature diversity present in the original dataset. The resampled set of top-$k$ features, $\mathcal{K'}$, is then used for sample selection. Importantly, our final dataset is used for training \textit{with the data order preserved}. Thus, FtF trains on samples in descending order of importance.

In our experiments we set $\alpha = 0.2$ as an intuitive value that balances importance preservation with increasing diversity. Future work can explore optimizing $\alpha$.

\section{Experiments}
\label{experiments}

\paragraph{Filter Pair} 
We use Gemma 3 12B IT \cite{team_gemma_2025} and its layer 24 residual stream SAE with a 65k width and "medium" L0 \cite{mcdougall_gemma_2025}.

\paragraph{Models and Training}
We perform SFT with LoRA parameter-efficient finetuning \cite{hu_lora_2021}. We train 2 different model-dataset pairs in our main experiments: 
\begin{enumerate}[nosep]
    \item Llama 3.1 8B \cite{grattafiori_llama_2024} on UltraChat 200k \cite{ding_enhancing_2023}
    \item Qwen 3 8B (Base) \cite{team_qwen3_2025} on Alpaca (Cleaned) \cite{taori_alpaca_2023}
\end{enumerate}
These represent a diversity of model families and data sources, providing a test of generalization. Following our sweeps for parameter selection\footnote{see Appendix \ref{a:param-select} for details on our parameter selection process}, we set $k=5$ and $N=100$ in our experiments. To ensure that no method is simply taking advantage of the benefits of longer sequences \cite{zhao_long_2024} to select samples, we use a fixed sequence length of 256 tokens. Further details on training parameters are provided in Appendix \ref{a:train-params}.

\paragraph{Comparisons}
We present comparisons with a reference of randomly selected subsets at the $N=100$ dataset size, a referece of training on the entire dataset (\textbf{All}), and 2 SOTA comparisons: GRAPE \cite{zhang_best_2025} and S2L \cite{yang_smalltolarge_2024}.

\paragraph{Evaluation}
We measure the quality of each trained model across two paradigms of evaluation: traditional benchmarking with Open LLM Leaderboard v2 \cite{fourrier_open_2024}, and LLM-as-a-judge with AlpacaEval 2.0\footnote{To reduce costs, we run AlpacaEval 2.0 locally using Qwen 3.8 27B FP8 as our judge} \cite{li_alpacaeval_2023}. We additionally include the harmonic mean of the seed-averaged Leaderboard and AlpacaEval scores (HMean) as a single metric which requires high scores in both evaluation paradigms to produce a high composite score. Motivated by \citet{polo_tinybenchmarks_2024}\footnote{Their key finding was that only 100 samples can approximate full evaluation performance on a selection of benchmarks, including AlpacaEval 2.0}, we perform a more efficient evaluation with AlpacaEval 2.0 by using a 300-sample subset. Further details on evaluation parameters are provided in Appendix \ref{a:eval-params}

\paragraph{Hardware}
We run all of our experiments on a single server with 2x NVIDIA RTX A6000 48GB GPUs.

\section{Main Results}
\label{s:main-results}

\subsection{Task Performance}
All results are averages across 3 seeds \{6, 7, 67\}. 
Appendix \ref{a:param-select} contains the results for the parameter sweeps we conducted to determine the $k$ value of 5 at the $N=100$ training budget. Table \ref{tab:llama31-ultrachat-summary} contains the results summary for Llama 3.1 8B trained on UltraChat 200k. Table \ref{tab:qwen3-alpaca-summary} contains the summary for Qwen 3 8B (Base) trained on Alpaca (cleaned). Detailed Leaderboard results can be found for Llama/UltraChat in Table \ref{tab:llama31-ultrachat-detailed} and for Qwen/Alpaca in Table \ref{tab:qwen3-alpaca-detailed}.

Our experiments demonstrate just how challenging generalization is. Neither FtF, GRAPE, nor S2L claims a decisive victory, but all show similar levels of success overall. Of these 3, GRAPE performs worst in the Llama/UltraChat experiments, but best in Qwen/Alpaca; the opposite is true for FtF. S2L performs second-best in both sets of experiments, a notable point of consistency. Interestingly, the results for Llama/UltraChat are consistently much lower than Qwen/Alpaca. FtF outperforms GRAPE by over 230\% in Llama/UltraChat HMean, corresponding to 1.28 percentage points. In Qwen/Alpaca, the absolute difference is larger at 2.1 HMean percentage points, though that represents only a 10.2\% advantage. 

These results indicate that FtF is competitive with the current SOTA methods, establishing itself as similarly performant across a diversity of experiments.

\begin{table}[tbp]
    \centering
    \begin{tabular}{lccc}
        \toprule
        Model & AlpacaEval & Leaderboard & HMean \\
        \midrule
        Ref (\textbf{All}) & 0.31 & \textbf{26.85} & 0.61 \\
        Ref (100) & 0.00 & 24.13 & 0.00 \\
        GRAPE & 0.28 & 23.85 & 0.55 \\
        S2L & 0.55 & 23.99 & 1.07 \\
        FtF & \textbf{0.95} & 23.70 & \textbf{1.83} \\
        \bottomrule
    \end{tabular}
    \caption{Task performance results for Llama 3.1 8B trained on UltraChat 200k.
    Bold indicates the best result in each column.}
    \label{tab:llama31-ultrachat-summary}
\end{table}

\begin{table}[tbp]
    \centering
    \begin{tabular}{lccc}
        \toprule
        Model & AlpacaEval & Leaderboard & HMean \\
        \midrule
        Ref (\textbf{All}) & 12.16 & \textbf{42.58} & 18.92 \\
        Ref (100) & 12.00 & 40.80 & 18.55 \\
        GRAPE & \textbf{15.60} & 41.10 & \textbf{22.61} \\
        S2L & 14.77 & 41.07 & 21.73 \\
        FtF & 13.66 & 41.17 & 20.51 \\
        \bottomrule
    \end{tabular}
    \caption{Task performance results for Qwen 3 8B (Base) trained on Alpaca (cleaned).
    Bold indicates the best result in each column.}
    \label{tab:qwen3-alpaca-summary}
\end{table}

\subsection{Top-$k$ Features}
\label{ss:top-k-feats}
Here we present the ranks, indices, and descriptions of the most important SAE features identified by FtF for both sets of experiments. 
Feature descriptions were collected via the Neuronpedia \cite{lin_neuronpedia_2023} interactive database for Gemma Scope 2 SAEs. 

With our diversity resampling parameter $\alpha=0.2$, the top-4 features are the "true" top-4, with the fifth being the resampled diversity feature. Interestingly, we found that the same top-4 features were identified by both the Llama/UltraChat and Qwen/Alpaca sets of experiments. Table \ref{tab:feature_rankings} contains the details for those top-4 features.

\begin{table}[htbp]
    \centering
    \begin{tabular}{ccp{0.65\linewidth}}
        \toprule
        Rank & Index & Description \\
        \midrule
         1 & 1198 & possessive 's \\
         2 & 480 & numbers after quotes \\
         3 & 329 & \textit{not available} \\
         4 & 1067 & quotation marks followed by numbers \\
        \bottomrule
    \end{tabular}
    \caption{Top-$k$ features identified by FtF for both Llama/UltraChat and Qwen/Alpaca sets of experiments.}
    \label{tab:feature_rankings}
\end{table}

Unfortunately, the current state of open-source descriptions for our features leaves much to be desired. For features 1198, 329, and 1067, there were no records of token activations available, meaning that there is no evidence to support or verify that the available feature description is accurate. Feature 329 had no available explanation. Feature 480 was the only of the top-4 features which had both an explanation and top activating examples available. However, those top activating examples all indicate high activations for spaces before numbers, as well as the apostrophes in "Let's" and "It's"-- not necessarily consistent with the available feature description. Neuronpedia does have a feature auto-interpretability tool, but the cost of generating feature descriptions is a notable accessibility barrier for feature interpretability.

\section{Analysis}
\label{s:analysis}

In this section we provide analyses for different components of FtF, as well as for feature generalization. Here we use the same model--dataset pair used in our parameter sweeps in Appendix \ref{a:param-select} (i.e. SmolLM3 3B (Base) and Alpaca (Cleaned)).  As with all our other FtF experiments, we set $k=5$ and $N=100$.

\subsection{Learning Rate Scheduler}
\label{ss:lr}
FtF trains on selected samples in descending order of their feature-based importance scores. A decaying learning rate therefore places greater optimization weight on earlier, higher-ranked samples. To examine whether FtF's advantage depends on this interaction, we compare cosine and constant learning rate schedules. FtF outperforms the random reference under both schedules, with HMean improvements of 1.41 and 1.12 percentage points, respectively. The larger improvement under cosine scheduling is consistent with the hypothesis that emphasizing higher-ranked samples is beneficial. We interpret these results as evidence that FtF's selection benefit persists across schedules, with suggestive evidence of an additional interaction between importance ordering and learning-rate decay.

\begin{table}[htbp]
    \centering
    \begin{tabular}{lccc}
        \toprule
        Model & AlpacaEval & Leaderboard & HMean \\
        \midrule
        Ref (constant) & 1.92 & 30.60 & 3.62 \\
        Constant & \textbf{2.55} & \textbf{32.28} & \textbf{4.74} \\
        \midrule
        Ref (cosine) & 1.45 & 31.50 & 2.77 \\
        Cosine & 2.24 & 31.50 & 4.18 \\
        \bottomrule
    \end{tabular}
    \caption{Comparison of constant and cosine learning rate schedulers
    for FtF with $k=5$ and $N=100$. Results are percentages. HMean is the
    harmonic mean of AlpacaEval and Open LLM Leaderboard scores.
    \textbf{Ref} randomly selects an $N$-sample dataset and uses the
    indicated learning rate scheduler.
    Bold indicates the highest score in each column.}
    \label{tab:lr-scheduler-hmean}
\end{table}

\subsection{Superfiltering}
\label{ss:superfiltering}

As discussed in the background in Section \ref{s:bg-rw}, \textit{superfiltering} is the general term for using small models to perform efficient data filtering and selection for training larger models. One of our SOTA comparisons, S2L, is an example of such a method. We explore how superfiltering performs in FtF by using a far smaller model (Gemma 3 270M, instead of Gemma 3 12B IT) and its corresponding SAE as the filter pair. Note that we \textbf{do not} use the instruction-tuned version of the small model, evaluating whether a very small model without specialized knowledge of the target task can serve as useful filter pair. 

Success in this scenario would indicate that FtF may be: (1) widely accessible, with small models reducing the hardware barrier; and (2) generalizable for different post-training tasks, not requiring a model with specialized task knowledge to enable efficient training for that specialized task. The second point is of particular importance, as there are currently only a handful of publicly-available SAEs. 

Results for the summary of the superfiltering analysis are in Table \ref{tab:superfiltering-summary}, and the detailed Leaderboard results are in Table \ref{tab:superfiltering-details}. Although the 270M filter pair fails to outperform the 12B IT variant, it performs 40.1\% better than the random reference and falls only 6.7\% short of the large filter pair. Furthermore, the LLM of the 12B IT filter pair (responsible for the vast majority of the parameters in the filter pair) is over 44 times larger than 270M model. For many scenarios, a 6.7\% performance deficit at only 2.3\% of the parameter count may be a useful and valuable tradeoff.

\begin{table}[htbp]
    \centering
    \begin{tabular}{lccc}
        \toprule
        Model & AlpacaEval & Leaderboard & HMean \\
        \midrule
        Reference & 1.45 & \textbf{31.50} & 2.77 \\
        270M PT & 2.09 & 30.53 & 3.90 \\
        12B IT & \textbf{2.24} & \textbf{31.50} & \textbf{4.18} \\
        \bottomrule
    \end{tabular}
    \caption{Superfiltering analysis summary. AlpacaEval, Open LLM Leaderboard, and harmonic mean results
    comparing 270M PT and 12B IT in FtF with $k=5$ and $N=100$. Reference is trained on a randomly selected $N=100$ sample dataset.
    Results are percentages.
    Bold indicates the highest score in each column.}
    \label{tab:superfiltering-summary}
\end{table}

\subsection{Sequence Aggregation}
\label{ss:seq-agg}
Aggregating feature activations across a full sample token sequence is an important consideration in assessing the semantic content of a training sample. In our main experiments, we apply last-token-pooling because the last token is the only position which attends to every token in the sequence, due to the causal masking. However, SFT trains at token-level, so a training loss is computed at each token position, and thus is trained on the features present in each causal subsequence. 

We perform an analysis to explore how mean-pooling (Mean) performs in comparison to last-token-pooling (LT). The summarized results are shown in Table \ref{tab:mean-pool-hmean}, and the detailed Leaderboard results are in Table \ref{tab:mean-lb-details}. Last-token-pooling demonstrates a small advantage over mean-pooling, scoring 0.5 percentage points higher in HMean, with the gap primarily from a 0.28 percentage point lead in AlpacaEval. Given this performance advantage in this analysis and the intuition behind evaluating the feature content only with the full sequence in context, last-token-pooling indeed appears to be the more reasonable selection.

\begin{table}[htbp]
    \centering
    \begin{tabular}{lccc}
        \toprule
        Model & AlpacaEval & Leaderboard & HMean \\
        \midrule
        Reference & 1.45 & \textbf{31.50} & 2.77 \\
        Mean & 1.96 & 31.46 & 3.68 \\
        LT & \textbf{2.24} & \textbf{31.50} & \textbf{4.18} \\
        \bottomrule
    \end{tabular}
    \caption{Sequence aggregation analysis summary. Comparison of last-token pooling (LT) and mean pooling (Mean) for FtF with $k=5$ and $N=100$. Results are percentages.
    Reference is trained on a randomly selected $N=100$ sample dataset.
    Bold indicates the highest score in each column,
    including ties at the displayed precision.}
    \label{tab:mean-pool-hmean}
\end{table}

\subsection{Ablation Studies}
\label{ss:ablations}

\subsubsection{Diversity Resampling}
In FtF, we willingly sacrifice a singular focus on the most important features for less important ones in an attempt to promote feature diversity and more complete coverage of the feature distribution present in the original dataset. Here we analyze whether this sacrifice does indeed lead to better model performance.

In Table \ref{tab:divoff-hmean} we present the task performance summaries of running FtF with diversity resampling (Div)-- exactly as in our main experiments-- and without resampling the features (No Div). Table \ref{tab:divoff-lb-details} contains the detailed Leaderboard results. Our ablation analysis indicates that the importance sacrifice was indeed worth it, with Div outperforming No Div in every metric except MuSR. Our experiments were run with $k=5$ and $\alpha=0.2$, which translates to resampling a single feature. Future work may explore how to tune $\alpha$ across different values of $k$.

\begin{table}[htbp]
    \centering
    \begin{tabular}{lccc}
        \toprule
        Model & AlpacaEval & Leaderboard & HMean \\
        \midrule
        Reference & 1.45 & \textbf{31.50} & 2.77 \\
        No Div & 1.96 & 31.44 & 3.68 \\
        Div & \textbf{2.24} & \textbf{31.50} & \textbf{4.18} \\
        \bottomrule
    \end{tabular}
    \caption{Diversity resampling ablation summary. Comparison of FtF with and without feature diversity resampling. Results are percentages.
    Reference is trained on a randomly selected $N=100$ sample dataset.
    Bold indicates the highest score in each column,
    including ties at the displayed precision.}
    \label{tab:divoff-hmean}
\end{table}

\subsubsection{Feature Entropy}
As shown in Equation \ref{eq:importance_w_ent}, the conditional entropy of active features plays a central role in determining the importance of a feature. Its inclusion in our feature importance was influenced by the findings of \citet{zhang_best_2025} and their GRAPE methodology: training on data which is easier for the model to learn from appears to produce higher quality models. We applied that philosophy to feature importance in FtF and arrived at conditional entropy being our heuristic for a notion of "ease of learning".

In this ablation study, we analyze whether this heuristic actually adds any value to our methodology. We compare FtF with entropy included in the feature importance calculation (Ent), and without (No Ent). Table \ref{tab:no-ent-hmean} contains the summary of the study, and Table \ref{tab:no-ent-lb-details} contains the detailed Leaderboard task results. Here we see very similar results to our diversity resampling ablation, as well as the sequence aggregation analysis, with our formulation of FtF performing slightly better than the alternatives explored.

\begin{table}[htbp]
    \centering
    \begin{tabular}{lccc}
        \toprule
        Model & AlpacaEval & Leaderboard & HMean \\
        \midrule
        Reference & 1.45 & \textbf{31.50} & 2.77 \\
        No Ent & 1.96 & 31.46 & 3.68 \\
        Ent & \textbf{2.24} & \textbf{31.50} & \textbf{4.18} \\
        \bottomrule
    \end{tabular}
    \caption{Feature entropy ablation summary. Comparison of FtF with and without entropy included in our definition of feature importance.
    Reference is trained on a randomly selected $N=100$ sample dataset.
    Bold indicates the highest score in each column,
    including ties at the displayed precision.}
    \label{tab:no-ent-hmean}
\end{table}

\subsection{Feature Importance Across Datasets}
\label{ss:feature-consistency}
After seeing consistency in the top-4 features identified by FtF in Section \ref{ss:top-k-feats}, we ran a deeper analysis of the feature ranking consistency across the two datasets. In Table \ref{tab:feature-ranking-correlations} 
we present the feature importance rank correlations across the 2 datasets. We run correlations for the full 65k feature ranking, as well as for a selection of subsets of the top features.

In general there is a high degree of agreement between these two datasets, including perfect alignment for the $k=5$ and $k=10$ subsets, and an overall correlation of $\rho=0.9073$. As expected, larger subsets gradually reduce the amount of overlap, with the negative correlations being an artifact of our decision to assign absent features a rank of $k+1$. Interestingly, we observe somewhat of a recovery, with $k=50$ exhibiting a similar frcation of shared features as $k=1000$. We interpret this as evidence to suggest that there may be general consistency in the features associated with instruction-following.

Consistency and similarity in important features across datasets can serve to amortize the cost of identifying the top-$k$ features by enabling the reuse of feature importance rankings. Future work may further explore the potential universality of important features for instruction-following and other model behaviors.

\begin{table}[htbp]
    \centering
    \begin{tabular}{rccc}
        \toprule
        $k$ & Spearman's $\rho$ & Union size & Fraction shared \\
        \midrule
        5     & $1.0000$  & 5      & 100.0\% \\
        10    & $1.0000$  & 10     & 100.0\% \\
        50    & $0.6258$  & 67     & 66.0\% \\
        100   & $0.2103$  & 155    & 45.0\% \\
        200   & $-0.1861$ & 347    & 26.5\% \\
        500   & $-0.4372$ & 793    & 41.4\% \\
        1000  & $-0.3133$ & 1,357  & 64.3\% \\
        All   & $0.9073$  & 65,536 & 100.0\% \\
        \bottomrule
    \end{tabular}
    \caption{Feature-ranking correlations between UltraChat 200K
    and Alpaca (Cleaned). Correlations are computed over
    the union of the two top-$k$ feature sets. A feature absent from one dataset's top-$k$ subset is assigned rank $k+1$. Fraction shared
    is the intersection size divided by $k$,
    $(2k-\lvert U\rvert)/k$, where $U$ is the union.}
    \label{tab:feature-ranking-correlations}
\end{table}

\section{Discussion}
\label{s:discussion}

\paragraph{Decoder Norms in Feature Importance}
Some SAEs, such as standard ReLU and Gated SAEs, require explicitly unit-normalizing the decoder vectors during training \cite{bricken_towards_2023, huben_sparse_2023, rajamanoharan_improving_2024} while others, such as TopK, BatchTopK, and JumpReLU, allow for training SAEs without a strict unit-norm constraint \cite{rajamanoharan_jumping_2024, gao_scaling_2024, bussmann_batchtopk_2024}. Our methodology for computing feature importance includes the decoder norms in its formulation for compatibility with all SAE variants. Many SAEs, including the Gemma Scope 2 SAEs \cite{mcdougall_gemma_2025} that we use in our experiments, are unit-normalized, even when not mathematically necessary.

\paragraph{Compositional Methodologies}
One interesting direction for future work is an exploration of how different data-efficient IT methods may be combined into a compositional methodology. For example, FtF can be used in conjunction with a method such as GRAPE to produce a filtered dataset which captures the most important features and contains samples easiest for the model to learn from. 

\paragraph{Methodological Cost Comparison}
Although instruction-tuning is a one-time cost, it is worth considering how FtF compares to our SOTA comparisons in the costs associated with each methodology. Similar to GRAPE, the main costs of FtF are inference over the dataset. In our experiments, FtF took around 6.5 hours to perform data selection on UltraChat 200K, whereas GRAPE took about 14.5. Furthermore, GRAPE was only run using the existing dataset, and we elected not to use the version of GRAPE which generates and scores candidate responses from a selection of different LLMs, as that would have made our GRAPE experiments infeasible to run. The main costs associated with S2L are training the proxy model for multiple passes over the entire dataset. Although small proxy models help reduce the costs of S2L, the resource requirements for training a model for multiple full passes of a dataset are much higher than the inference-based methods.

\paragraph{Practical Utility of Feature Interpretability}
One of the key advantages of FtF over other SOTA methodologies such as GRAPE and S2L is its focus on feature interpretability. While interpretability does not inherently produce practical benefits, there are applications where a semantic understanding of data is essential. One application is in identifying coverage gaps in a dataset to inform precise collection and curation of additional data \cite{wang_data_2024, liang_aligning_2025}. Another application is building connections and understanding the relationships between training data and resulting model behavior \cite{lu_instag_2024, dunlap_describing_2024, li_where_2025}. FtF is also uniquely positioned to enable human auditing and intervention, similar to SHIFT \cite{marks_sparse_2025}, although the quality and accessibility issues of feature explanations described in Section \ref{ss:top-k-feats} are currently a barrier to further exploring this.

\paragraph{Limitations}
There are a few limitations of this work we aim to address. The first is the lack of exploration of different SAE variants in the filter pairs. Varying the LLM layer, feature width, and L0 values of the SAE can change the nature of the features identified by FtF and may be an interesting direction for future work. The second limitation is the size of the models in our experiments and our focus on a single data budget. Constraints in time/compute meant that we had to limit the scope of our experiments, and we felt that the 8B parameter scale and $N=100$ data budget presented a compelling case for FtF's utility while being efficient enough for the resources at our disposal. The final limitation is the lack of an example of human-in-the-loop intervention with FtF. This work was heavily inspired by \citet{marks_sparse_2025} and their SHIFT methodology, and it is unfortunate that we did not have the resources available to perform a similar intervention with FtF.

\section{Conclusion}
\label{s:conclusion}
We introduced Follow the Features (FtF), an instruction-tuning data-selection method that uses SAE activation statistics to identify prominent features in a dataset and select samples that strongly express them. By separating the filter model from the training model, FtF enables feature-based selection across target model families. At a training budget of 100 samples with a fixed sequence length of 256 tokens, FtF achieves performance competitive with GRAPE and S2L across two model–dataset settings. Its selected subsets outperform full-dataset training on the reported harmonic-mean metric, while full-dataset training achieves higher Open LLM Leaderboard scores.

Our analyses also suggest that a substantially smaller filter model can retain much of the selection benefit in the evaluated setting. However, the semantic meaning of the highest-ranked features remains incompletely verified, and the practical value of feature-guided human intervention has yet to be demonstrated. Future work should examine these features directly, evaluate selection across broader training budgets, and measure the end-to-end cost of filtering and fine-tuning. FtF provides a foundation for investigating how explicit feature representations can support both effective data selection and more transparent dataset curation.

\section*{AI Usage Disclosure}
We used generative AI tools for suggesting training and evaluation parameters, creating and editing code, summarizing existing literature, identifying research gaps, brainstorming different entropy variants, identifying relevant literature, formatting references, creating \LaTeX tables and Ti\emph{k}Z figures, and editing the paper. OpenCode was used as the agentic coding harness, and ChatGPT was used for literature and information retrieval, parameter suggestions, brainstorming, and making final edits to the manuscript.

\clearpage


\bibliography{references}
\bibliographystyle{mlsys2025}

\clearpage

\appendix

\section{Feature Entropy Implementation Details}
\label{a:feat-entropy-details}

We first compute a Gaussian approximation to the differential entropy of each
feature:
\begin{equation}
    H_{\mathrm{Gauss}}(X)
    = \frac{1}{2}\ln(2\pi e\sigma^2),
    \label{eq:univariate_gaussian_entropy}
\end{equation}
where entropy is measured in nats. The mean and variance are computed from
positive SAE activations over all token positions, without sequence pooling.
The resulting coarse importance score is
\begin{equation}
    I_j^{\mathrm{coarse}}
    =
    \frac{\mu_j \lVert w_j^{\mathrm{dec}}\rVert_2}
    {\max(H_{\mathrm{Gauss},j},10^{-8})}.
\end{equation}

Only the 1,000 features with the highest coarse importance are retained for
entropy refinement. For each retained feature, we collect at most 2,048
positive activations and estimate differential entropy using
\texttt{scipy.stats.differential\_entropy}. Before estimation, independent
uniform noise in $[-10^{-8},10^{-8}]$ is added to break ties. The refined
entropy divisor is floored at $0.01$.

Features with at most 20 positive activations are not entropy-refined and use
an entropy divisor of $1$. Dead features have zero positive activations and
therefore zero mean importance; no separate correction is applied to them.
Constant or tied activations are handled only through the small jitter
described above.

\section{FtF Parameter Selection}
\label{a:param-select}
We ran our sweeps with SmolLM3 3B (Base) \cite{bakouch_smollm3_2025}-- a model which balances high capability with a compact, efficient size-- and the Alpaca (Cleaned) dataset \cite{taori_alpaca_2023}. We fixed the parameters set using the results of these sweeps and applied them to our main experiments described in Section \ref{experiments}. Thus, the results in Section \ref{s:main-results} reflect parameter generalization, as well as overall methodology generalization. 

We searched for $k$ values in \{1, 5, 10, 100, 200\} and used data budgets of \{100, 1,000, 10,000\}. 200 was chosen for the highest $k$ value because we believe that to be a reasonable value for the upper bound of features that a human might review and audit. 10,000 was selected as the largest data budget following prior work done by \citet{liu_what_2024}, which established that 10,000 samples was a useful reduction for Alpaca. We chose to apply the fixed data budget, rather than relative to the size of the original dataset, as a realistic compute budget would be restricted by a fixed total number of training samples, not by a fixed percentage of the entire dataset.

To increase the efficiency of our sweep, we separated it into 2 phases: Phase 1 is a full pass over every parameter combination for a single seed, and Phase 2 is a pass over the top 2 performing $k$ values at each training budget for the full 3 seed sweep. 

\subsection{Phase 1: Wide, Single Seed}
Since we only run Phase 1 for a single seed \{6\}, we take the top 2 $k$ values at each dataset size for a targeted sweep in Phase 2. From Table \ref{tab:ftff-sweep-hmean-p1} we see that at $N=100$ and $N=1,000$, $k=1$ and $k=5$ are the best performers and achieve virtually identical HMean results, including lifts of 2.59 and 1.21 over the random Reference at $N=100$ and $N=1,000$, respectively. At $N=10,000$, the separation between the Reference and best FtF variant is much tighter at only 0.34 for HMean at $k=200$. While these Phase 1 results are indicative of promising parameters, Phase 2 further evaluates the robustness of the best Phase 1 results.

\begin{table}[htbp]
    \centering
    \begin{tabular}{lccc}
        \toprule
        Model & AlpacaEval & Leaderboard & HMean \\
        \midrule
        Baseline & 0.79 & 28.88 & 1.55 \\
        Ref (All) & 2.45 & 32.29 & 4.55 \\
        \midrule
        \multicolumn{4}{l}{\textbf{Dataset size: 100}} \\
        \addlinespace
        Reference & 1.45 & \textbf{31.50} & 2.77 \\
        $k=1$     & \textbf{\textit{2.93}} & 31.31 & \textbf{\textit{5.36}} \\
        $k=5$     & \textbf{\textit{2.93}} & 31.31 & \textbf{\textit{5.36}} \\
        $k=10$    & 2.11 & 30.53 & 3.94 \\
        $k=100$   & 1.44 & 31.05 & 2.75 \\
        $k=200$   & 2.11 & 31.05 & 3.95 \\
        \midrule
        \multicolumn{4}{l}{\textbf{Dataset size: 1,000}} \\
        \addlinespace
        Reference & 1.47 & 31.41 & 2.82 \\
        $k=1$     & \textbf{2.15} & 31.62 & \textbf{4.03} \\
        $k=5$     & \textbf{2.15} & \textbf{31.64} & \textbf{4.03} \\
        $k=10$    & 2.10 & 31.04 & 3.93 \\
        $k=100$   & 0.85 & 30.53 & 1.66 \\
        $k=200$   & 0.94 & 31.34 & 1.83 \\
        \midrule
        \multicolumn{4}{l}{\textbf{Dataset size: 10,000}} \\
        \addlinespace
        Reference & 2.37 & 32.18 & 4.42 \\
        $k=1$     & 1.02 & 32.14 & 1.97 \\
        $k=5$     & 1.02 & 32.14 & 1.97 \\
        $k=10$    & 0.97 & 31.87 & 1.88 \\
        $k=100$   & 2.40 & 32.15 & 4.47 \\
        $k=200$   & \textbf{2.57} & \textbf{\textit{32.61}} & \textbf{4.76} \\
        \bottomrule
    \end{tabular}
    \caption{Results for the sweep across dataset sizes and $k$ values
    for FtF Phase 1. Results are percentages. HMean is the harmonic mean
    of AlpacaEval and Open LLM Leaderboard scores.
    \textbf{Baseline} is the starting checkpoint (i.e., before training)
    and \textbf{Reference} trains on $N$ randomly selected data samples, with All indicating a random ordering of the full dataset.
    Bold indicates the highest score within each dataset size,
    including ties at the displayed precision. Italics indicate the best observed value at any dataset size.}
    \label{tab:ftff-sweep-hmean-p1}
\end{table}

\begin{table*}[tbp]
    \centering
    \begin{tabular}{lcccccc}
    \toprule
    Model & BBH & GPQA & IFEval & MATH-Hard & MMLU-Pro & MuSR \\
    \midrule
    Baseline
        & 42.27 & 24.92 & 26.14
        & \textbf{\textit{16.16}} & 31.37 & 32.41 \\
    Ref (All)
        & 45.39 & 27.29 & \textbf{\textit{41.41}}
        & 12.99 & 29.86 & 36.82 \\
    \midrule
    \multicolumn{7}{l}{\textbf{Dataset size: 100}} \\
    \addlinespace
    Reference
        & 43.03 & 27.01 & \textbf{35.01}
        & 15.11 & 31.28 & \textbf{\textit{37.57}} \\
    $k=1$
        & 43.13 & \textbf{\textit{28.52}} & \textbf{35.01}
        & \textbf{15.26} & 31.40 & 34.52 \\
    $k=5$
        & 43.13 & \textbf{\textit{28.52}} & \textbf{35.01}
        & \textbf{15.26} & 31.40 & 34.52 \\
    $k=10$
        & \textbf{43.46} & 26.76 & 31.65
        & 15.11 & 30.90 & 35.32 \\
    $k=100$
        & 43.17 & 27.35 & 34.29
        & 13.90 & 31.23 & 36.38 \\
    $k=200$
        & 43.03 & 26.68 & 33.33
        & 15.11 & \textbf{\textit{31.48}} & 36.64 \\
    \midrule
    \multicolumn{7}{l}{\textbf{Dataset size: 1,000}} \\
    \addlinespace
    Reference
        & \textbf{45.17} & 27.68 & \textbf{34.17}
        & 14.65 & 30.43 & \textbf{36.38} \\
    $k=1$
        & 44.59 & \textbf{28.02} & 33.69
        & \textbf{15.71} & \textbf{31.47} & 36.24 \\
    $k=5$
        & 44.59 & \textbf{28.02} & 33.81
        & \textbf{15.71} & \textbf{31.47} & 36.24 \\
    $k=10$
        & 44.32 & 27.18 & 32.73
        & 14.73 & 31.31 & 35.98 \\
    $k=100$
        & 45.10 & \textbf{28.02} & 29.74
        & 14.80 & 31.25 & 34.26 \\
    $k=200$
        & 44.71 & 27.85 & 34.05
        & 15.48 & 31.42 & 34.52 \\
    \midrule
    \multicolumn{7}{l}{\textbf{Dataset size: 10,000}} \\
    \addlinespace
    Reference
        & 45.44 & 27.01 & 38.97
        & 14.20 & 30.66 & 36.77 \\
    $k=1$
        & \textbf{\textit{45.76}} & 26.85 & 38.49
        & 13.67 & \textbf{31.15} & 36.90 \\
    $k=5$
        & \textbf{\textit{45.76}} & 26.85 & 38.49
        & 13.67 & \textbf{31.15} & 36.90 \\
    $k=10$
        & 45.17 & 27.01 & 38.97
        & 13.07 & 30.35 & 36.64 \\
    $k=100$
        & 45.25 & \textbf{27.43} & 38.61
        & 13.97 & 30.48 & \textbf{37.17} \\
    $k=200$
        & \textbf{\textit{45.76}} & 27.01 & \textbf{41.25}
        & \textbf{15.26} & 30.54 & 35.85 \\
    \bottomrule
\end{tabular}
    \caption{Detailed Open LLM Leaderboard results by task for the
    FtF Phase 1 sweep, grouped by dataset size. Results are percentages.
    Bold indicates the highest score within each dataset size,
    including ties at the displayed precision. Italics indicate the best observed value at any dataset size.}
    \label{tab:ftff-sweep-lb-details-p1}
\end{table*}

\subsection{Phase 2: Targeted, All Seeds}
The Phase 2 results in Table \ref{tab:ftff-sweep-hmean-p2} paint a much less clear picture than Phase 1. At $N=100$, we see results similar to Phase 1, with both $k$ values achieving near-identical scores at least 1.4 higher in HMean than the Reference. However, there is almost no separation among the models trained at the $N=1,000$ data budget, and at $N=10,000$ the Reference outperforms each FtF model. For $N=100$ and $N=1,000$, we set $k=5$; even though $k=1$ demonstrates near-identical performance, we expect the complete absence of feature diversity to lead to poor generalization. For $N=10,000$, $k=200$ trades a 0.34 percentage point AlpacaEval deficit for a 0.26 percentage point advantage in Leaderboard, relative to the Reference. The lower scale results of AlpacaEval compared to the Leaderscores scores mean that HMean favors the Reference, despite having very close averages: 17.275 (Reference) vs. 17.235 ($k=200$). 

We suspect that the diversity resampling parameter $alpha=0.2$ is the cause of the performance degradation as $N$ increases. It is likely too large for the larger datasets (e.g. this represents $40$ important features being replaced with less important ones at $k=200$).
Given the results of our sweep, we run our main experiments with $k=5$ and $N=100$.

\begin{table}[htbp]
    \centering
    \begin{tabular}{lccc}
        \toprule
        Model & AlpacaEval & Leaderboard & HMean \\
        \midrule
        Baseline & 0.79 & 28.88 & 1.55 \\
        Ref (All) & \textbf{\textit{2.45}} & 32.29 & \textbf{\textit{4.55}} \\
        \midrule
        \multicolumn{4}{l}{\textbf{Dataset size: 100}} \\
        \addlinespace
        Reference & 1.45 & \textbf{31.50} & 2.77 \\
        $k=1$     & 2.23 & \textbf{31.50} & 4.17 \\
        $k=5$     & \textbf{2.24} & \textbf{31.50} & \textbf{4.18} \\
        \midrule
        \multicolumn{4}{l}{\textbf{Dataset size: 1,000}} \\
        \addlinespace
        Reference & 1.47 & 31.41 & 2.82 \\
        $k=1$     & \textbf{1.50} & 31.43 & \textbf{2.87} \\
        $k=5$     & \textbf{1.50} & \textbf{31.45} & \textbf{2.87} \\
        \midrule
        \multicolumn{4}{l}{\textbf{Dataset size: 10,000}} \\
        \addlinespace
        Reference & \textbf{2.37} & 32.18 & \textbf{4.42} \\
        $k=100$   & 1.68 & 31.76 & 3.18 \\
        $k=200$   & 2.03 & \textbf{\textit{32.44}} & 3.82 \\
        \bottomrule
    \end{tabular}
    \caption{Results for the sweep across dataset sizes and $k$ values
    for FtF Phase 2. Results are percentages. HMean is the harmonic mean
    of AlpacaEval and Open LLM Leaderboard scores.
    \textbf{Baseline} is the starting checkpoint (i.e., before training)
    and \textbf{Reference} trains on $N$ randomly selected data samples, with All indicating a random ordering of the full dataset.
    Bold indicates the highest score within each dataset size,
    including ties at the displayed precision. Italics indicate the best observed value at any dataset size.}
    \label{tab:ftff-sweep-hmean-p2}
\end{table}

\begin{table*}[tbp]
    \centering
    \begin{tabular}{lcccccc}
    \toprule
    Model & BBH & GPQA & IFEval & MATH-Hard & MMLU-Pro & MuSR \\
    \midrule
    Baseline
        & 42.27 & 24.92 & 26.14
        & \textbf{\textit{16.16}} & 31.37 & 32.41 \\
    Ref (All)
        & 45.39 & 27.29 & \textbf{\textit{41.41}}
        & 12.99 & 29.86 & 36.82 \\
    \midrule
    \multicolumn{7}{l}{\textbf{Dataset size: 100}} \\
    \addlinespace
    Reference
        & 43.03 & 27.01 & 35.01
        & \textbf{15.11} & 31.28 & \textbf{\textit{37.57}} \\
    $k=1$
        & 43.29 & 28.33 & \textbf{35.77}
        & 15.06 & 31.42 & 35.14 \\
    $k=5$
        & \textbf{43.39} & \textbf{\textit{28.41}} & 35.73
        & 14.95 & \textbf{\textit{31.44}} & 35.10 \\
    \midrule
    \multicolumn{7}{l}{\textbf{Dataset size: 1,000}} \\
    \addlinespace
    Reference
        & \textbf{45.17} & 27.68 & \textbf{34.17}
        & 14.65 & 30.43 & 36.38 \\
    $k=1$
        & 44.92 & \textbf{27.80} & 32.97
        & \textbf{15.06} & \textbf{31.26} & \textbf{36.60} \\
    $k=5$
        & 44.92 & \textbf{27.80} & 33.05
        & \textbf{15.06} & \textbf{31.26} & \textbf{36.60} \\
    \midrule
    \multicolumn{7}{l}{\textbf{Dataset size: 10,000}} \\
    \addlinespace
    Reference
        & 45.44 & 27.01 & 38.97
        & 14.20 & 30.66 & \textbf{36.77} \\
    $k=100$
        & 45.36 & 27.04 & 37.41
        & 14.43 & 30.63 & 35.71 \\
    $k=200$
        & \textbf{\textit{45.73}} & \textbf{27.24} & \textbf{39.45}
        & \textbf{15.18} & \textbf{31.09} & 35.98 \\
    \bottomrule
\end{tabular}
    \caption{Detailed Open LLM Leaderboard results by task for the
    FtF Phase 2 sweep, grouped by dataset size. Results are percentages.
    Bold indicates the highest score within each dataset size,
    including ties at the displayed precision. Italics indicate the best observed value at any dataset size.}
    \label{tab:ftff-sweep-lb-details-p2}
\end{table*}

\section{Training Parameters}
\label{a:train-params}

\begin{table}[H]
    \centering
    \resizebox{\columnwidth}{!}{%
        \begin{tabular}{ll}
            \hline
            \textbf{Parameter} & \textbf{Value} \\
            \hline
            Epochs & 1 \\
            Sequence length & 256 \\
            Per-device batch size & 32 \\
            Optimizer & \texttt{adamw\_torch\_fused} \\
            Learning rate & $2 \times 10^{-4}$ \\
            Scheduler & Cosine \\
            Weight decay & 0.01 \\
            Max grad norm & 0.3 \\
            Warmup schedule & 3\% \\
            Precision & \texttt{bf16} \\
            \hline
        \end{tabular}%
    }
    \caption{Supervised fine-tuning configuration.}
    \label{tab:sft-config}
\end{table}

\subsection{LoRA Parameters}
\begin{table}[H]
    \centering
    \resizebox{\columnwidth}{!}{%
        \begin{tabular}{ll}
            \hline
            \textbf{Parameter} & \textbf{Value} \\
            \hline
            Rank $r$ & 16 \\
            LoRA $\alpha$ & 32 \\
            LoRA dropout & 0.05 \\
            Bias & \texttt{none} \\
            Target modules & \texttt{q\_proj, k\_proj}, \\                    & \texttt{v\_proj, o\_proj}, \\
                           & \texttt{gate\_proj, up\_proj}, \\ & \texttt{down\_proj} \\
            \hline
        \end{tabular}
    }
    \caption{LoRA configuration.}
    \label{tab:lora-config}
\end{table}

\section{Evaluation Parameters}
\label{a:eval-params}

\begin{table}[H]
    \centering
    \resizebox{\columnwidth}{!}{%
        \begin{tabular}{ll}
            \hline
            \textbf{Parameter} & \textbf{Value} \\
            \hline
            Judge model & \texttt{Qwen 3.8 27B FP8} \\
            Dataset & \texttt{alpaca\_eval\_gpt4\_baseline} \\
            Dataset source & \texttt{tatsu-lab/alpaca\_eval} \\
            AlpacaEval samples & 300 \\
            Leaderboard samples & all \\
            Batch size & 4 (1 for Full FT) \\
            Candidate maximum new tokens & 256 \\
            Judge maximum tokens & 512 \\
            Candidate temperature & 0.0 \\
            Judge temperature & 0.0 \\
            Top-$p$ & 1.0 \\
            Judge thinking & Disabled \\
            \hline
        \end{tabular}%
    }
    \caption{Evaluation configuration.}
    \label{tab:eval-config}
\end{table}

\section{Additional Tables and Figures}

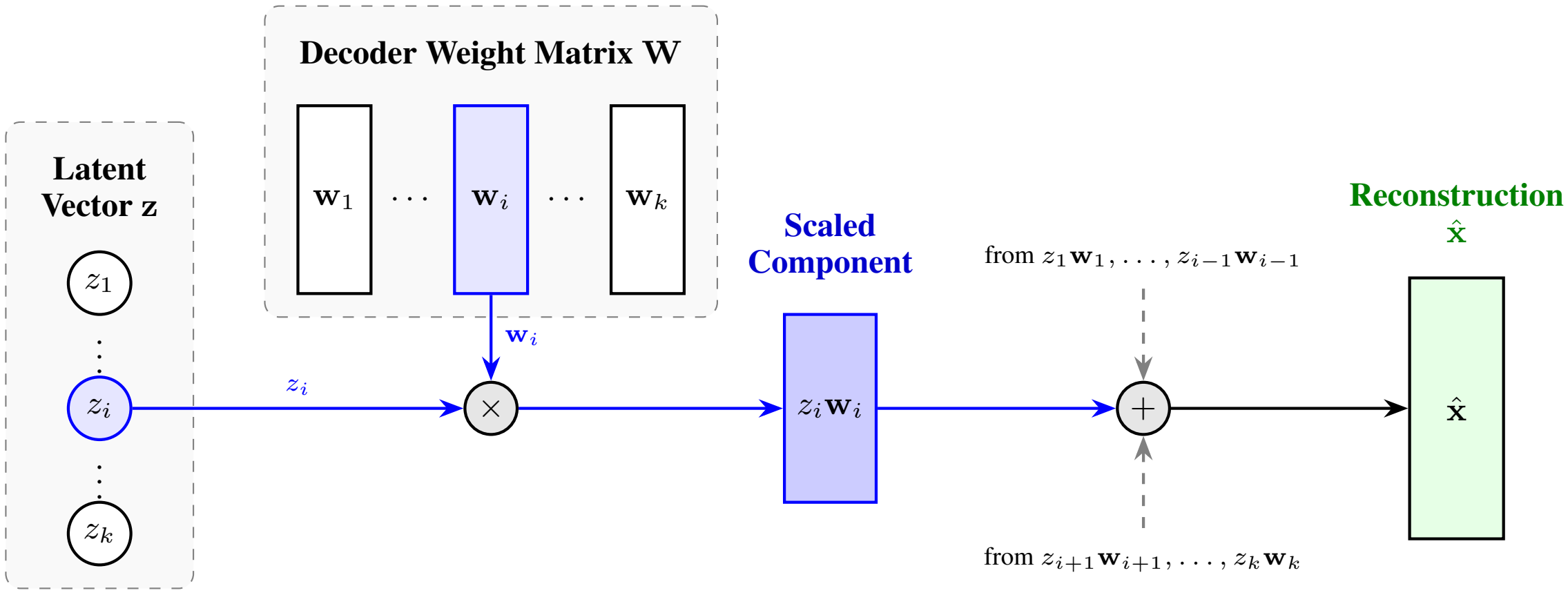
\begin{figure*}[h]
    \centering
    \begin{tikzpicture}[
        >=Stealth,
        neuron/.style={circle, draw=black, thick, minimum size=0.6cm, inner sep=0pt, font=\footnotesize, fill=white},
        hl_neuron/.style={circle, draw=blue, thick, fill=blue!10, minimum size=0.6cm, inner sep=0pt, font=\footnotesize\bfseries},
        vector/.style={rectangle, draw=black, thick, minimum width=0.7cm, minimum height=1.8cm, fill=white, align=center, font=\footnotesize},
        hl_vector/.style={rectangle, draw=blue, thick, minimum width=0.7cm, minimum height=1.8cm, fill=blue!10, font=\footnotesize\bfseries, align=center},
        operator/.style={circle, draw=black, thick, fill=gray!20, minimum size=0.5cm, inner sep=0pt, font=\small}
    ]
    
    \node[neuron] (z1) at (0, 1.2) {$z_1$};
    \node[font=\small] (vdots1) at (0, 0.6) {$\vdots$};
    \node[hl_neuron] (zi) at (0, 0) {$z_i$};
    \node[font=\small] (vdots2) at (0, -0.6) {$\vdots$};
    \node[neuron] (zk) at (0, -1.2) {$z_k$};
    
    \node[above=0.2cm of z1, font=\footnotesize\bfseries, align=center] (z_label) {Latent \\ Vector $\mathbf{z}$};
    
    \begin{scope}[on background layer]
        \node[fill=gray!5, draw=black!50, dashed, rounded corners, fit=(z1) (zk) (z_label), inner sep=6pt] (z_box) {};
    \end{scope}
    
    \def\xcoord{3.75}
    \node[vector] (w1) at ({\xcoord - 1.5}, 2.0) {$\mathbf{w}_1$};
    \node[font=\small] (vdotsw1) at ({\xcoord - 0.75}, 2.0) {$\cdots$};
    \node[hl_vector] (wi) at (\xcoord, 2.0) {$\mathbf{w}_i$};
    \node[font=\small] (vdotsw2) at ({\xcoord + 0.75}, 2.0) {$\cdots$};
    \node[vector] (wk) at ({\xcoord + 1.5}, 2.0) {$\mathbf{w}_k$};
    
    \node[above=0.2cm of wi, font=\footnotesize\bfseries, align=center] (w_label) {Decoder Weight Matrix $\mathbf{W}$};
    
    \begin{scope}[on background layer]
        \node[draw=black!50, dashed, rounded corners, fit=(w1) (wk) (w_label), inner sep=6pt, fill=gray!5] (w_box) {};
    \end{scope}
    
    \node[operator] (muli) at (\xcoord, 0) {$\times$};
    
    \draw[->, thick, blue] (zi) -- (muli) node[midway, above, font=\scriptsize, text=blue] {$z_i$};
    \draw[->, thick, blue] (wi) -- (muli) node[midway, right, font=\scriptsize, text=blue] {$\mathbf{w}_i$};
    
    \node[hl_vector, fill=blue!20] (ziwi) at (7.0, 0) {$z_i \mathbf{w}_i$};
    \node[above=0.2cm of ziwi, font=\footnotesize\bfseries, align=center, text=blue!80!black] {Scaled \\ Component};
    
    \draw[->, thick, blue] (muli) -- (ziwi);
    
    \node[operator] (sum) at (10.0, 0) {$+$};
    
    \draw[->, thick, blue] (ziwi) -- (sum);
    
    \draw[<-, thick, dashed, gray] (sum) -- ++(0, 1.2) node[above, text=black, font=\scriptsize] {from $z_1 \mathbf{w}_1, \dots, z_{i-1} \mathbf{w}_{i-1}$};
    \draw[<-, thick, dashed, gray] (sum) -- ++(0, -1.2) node[below, text=black, font=\scriptsize] {from $z_{i+1} \mathbf{w}_{i+1}, \dots, z_k \mathbf{w}_k$};
    
    \node[vector, fill=green!10, minimum height=2.5cm, minimum width=0.9cm] (xhat) at (13.0, 0) {$\hat{\mathbf{x}}$};
    \draw[->, thick] (sum) -- (xhat);
    \node[above=0.2cm of xhat, font=\footnotesize\bfseries, align=center, text=green!50!black] {Reconstruction \\ $\hat{\mathbf{x}}$};
    
    \end{tikzpicture}
    \vspace{-0.5em} 
    \caption{Illustration of the SAE decoder step. A specific activation $z_i$ in the latent space scales its corresponding column vector $\mathbf{w}_i$ in the decoder's weight matrix. The final reconstruction $\hat{\mathbf{x}}$ is the sum of all such scaled basis vectors.}
    \label{fig:autoencoder_latent_multiplication}
\end{figure*}

\begin{table*}[tbp]
    \centering
    \begin{tabular}{lcccccc}
        \toprule
        Model & BBH & GPQA & IFEval & MATH-Hard & MMLU-Pro & MuSR \\
        \midrule
        Ref (\textbf{All})
            & \textbf{39.36} & \textbf{27.63} & \textbf{29.62}
            & 3.70 & \textbf{23.85} & \textbf{36.95} \\
        Ref (100)
            & 37.85 & 25.06 & 26.30
            & \textbf{6.29} & 13.46 & 35.85 \\
        GRAPE
            & 37.51 & 25.03 & 26.54
            & 6.02 & 12.84 & 35.14 \\
        S2L
            & 37.57 & 24.97 & 26.30
            & 5.87 & 13.82 & 35.41 \\
        FtF
            & 37.10 & 25.06 & 26.10
            & 5.54 & 12.75 & 35.67 \\
        \bottomrule
    \end{tabular}
    \caption{Detailed Open LLM Leaderboard results for Llama 3.1 8B
    trained on UltraChat 200k.
    Bold indicates the best result in each column.}
    \label{tab:llama31-ultrachat-detailed}
\end{table*}

\begin{table*}[tbp]
    \centering
    \begin{tabular}{lcccccc}
        \toprule
        Model & BBH & GPQA & IFEval & MATH-Hard & MMLU-Pro & MuSR \\
        \midrule
        Ref (\textbf{All})
            & \textbf{58.77} & \textbf{35.23} & 47.16
            & 24.90 & 47.02 & \textbf{42.42} \\
        Ref (100)
            & 56.55 & 29.53 & 49.00
            & 24.30 & 47.39 & 38.05 \\
        GRAPE
            & 56.56 & 29.64 & \textbf{49.56}
            & \textbf{25.13} & 47.49 & 38.18 \\
        S2L
            & 56.78 & 29.81 & 49.00
            & 24.80 & 47.55 & 38.49 \\
        FtF
            & 56.51 & 29.53 & 49.52
            & 24.82 & \textbf{47.70} & 38.93 \\
        \bottomrule
    \end{tabular}
    \caption{Detailed Open LLM Leaderboard results for Qwen 3 8B (Base)
    trained on Alpaca (cleaned).
    Bold indicates the best result in each column.}
    \label{tab:qwen3-alpaca-detailed}
\end{table*}


\begin{table*}[tbp]
    \centering
    \begin{tabular}{lcccccc}
        \toprule
        Model & BBH & GPQA & IFEval & MATH-Hard & MMLU-Pro & MuSR \\
        \midrule
        Reference
            & 43.03 & 27.01 & 35.01
            & \textbf{15.11} & 31.28 & \textbf{37.57} \\
        Mean
            & 43.23 & 28.33 & 35.49
            & 14.90 & 31.41 & 35.36 \\
        LT
            & \textbf{43.39} & \textbf{28.41} & \textbf{35.73}
            & 14.95 & \textbf{31.44} & 35.10 \\
        \bottomrule
    \end{tabular}
    \caption{Mean sequence aggregation analysis details. Open LLM Leaderboard results by task comparing last-token pooling (LT) with mean pooling (Mean) for FtF with $k=5$ and $N=100$.
    Results are percentages.
    Bold indicates the highest score in each column,
    including ties at the displayed precision.}
    \label{tab:mean-lb-details}
\end{table*}

\begin{table*}[tbp]
    \centering
    \begin{tabular}{lcccccc}
        \toprule
        Model & BBH & GPQA & IFEval & MATH-Hard & MMLU-Pro & MuSR \\
        \midrule
        Reference
            & 43.03 & 27.01 & 35.01
            & \textbf{15.11} & 31.28 & \textbf{37.57} \\
        270M PT
            & 42.83 & 25.50 & 33.33
            & 14.12 & \textbf{31.48} & 35.93 \\
        12B IT
            & \textbf{43.39} & \textbf{28.41} & \textbf{35.73}
            & 14.95 & 31.44 & 35.10 \\
        \bottomrule
    \end{tabular}
    \caption{Superfiltering analysis. Detailed Open LLM Leaderboard results by task comparing
    270M PT and 12B IT for FtF with $k=5$ and $N=100$.
    Results are percentages.
    Bold indicates the highest score in each column.}
    \label{tab:superfiltering-details}
\end{table*}

\begin{table*}[tbp]
    \centering
    \begin{tabular}{lcccccc}
        \toprule
        Model & BBH & GPQA & IFEval & MATH-Hard & MMLU-Pro & MuSR \\
        \midrule
        Reference
            & 43.03 & 27.01 & 35.01
            & \textbf{15.11} & 31.28 & \textbf{37.57} \\
        No Div
            & 43.23 & 28.33 & 35.41
            & 14.90 & 31.41 & 35.36 \\
        Div
            & \textbf{43.39} & \textbf{28.41} & \textbf{35.73}
            & 14.95 & \textbf{31.44} & 35.10 \\
        \bottomrule
    \end{tabular}
    \caption{Diversity resampling ablation details. Open LLM Leaderboard results by task comparing FtF with (Div) and without (No Div) resampling for feature diversity.
    Results are percentages.
    Bold indicates the highest score in each column,
    including ties at the displayed precision.}
    \label{tab:divoff-lb-details}
\end{table*}

\begin{table*}[tbp]
    \centering
    \begin{tabular}{lcccccc}
        \toprule
        Model & BBH & GPQA & IFEval & MATH-Hard & MMLU-Pro & MuSR \\
        \midrule
        Reference
            & 43.03 & 27.01 & 35.01
            & \textbf{15.11} & 31.28 & \textbf{37.57} \\
        No Ent
            & 43.23 & 28.33 & 35.53
            & 14.90 & 31.41 & 35.36 \\
        Ent
            & \textbf{43.39} & \textbf{28.41} & \textbf{35.73}
            & 14.95 & \textbf{31.44} & 35.10 \\
        \bottomrule
    \end{tabular}
    \caption{Feature entropy ablation details. Open LLM Leaderboard results by task comparing FtF with (Ent) and without (No Ent) entropy in the feature importance definition.
    Bold indicates the highest score in each column,
    including ties at the displayed precision.}
    \label{tab:no-ent-lb-details}
\end{table*}

\end{document}